\documentclass[letterpaper, 10 pt, conference]{ieeeconf}  

\IEEEoverridecommandlockouts                              

\usepackage{graphicx}
\usepackage{color,soul}
\usepackage{hyperref}
\usepackage{mwe}
\usepackage{subcaption}
\usepackage{amsmath}
\usepackage{tabularx} 
\usepackage{multirow}

\newcolumntype{L}{>{\centering\arraybackslash}m{3cm}}
\usepackage{float}
\usepackage[thinc]{esdiff}
\usepackage{graphics} 
\usepackage[table]{xcolor}
\usepackage{caption}
\usepackage{graphicx}

\usepackage{graphics} 
\usepackage{epsfig} 
\usepackage{mathptmx} 
\usepackage{times} 
\usepackage{amsmath} 
\usepackage{amssymb}  
\usepackage[noadjust]{cite}
\usepackage{booktabs}
\usepackage{multirow}
\usepackage{algorithmic}
\usepackage{ragged2e}
\usepackage{array}

\newcolumntype{P}[1]{>{\RaggedRight\arraybackslash}p{#1}}
\newcolumntype{M}[1]{>{\centering\arraybackslash}m{#1}}

\usepackage{microtype}
\renewcommand{\baselinestretch}{1.00}

\usepackage[table]{xcolor}

\title{\Large \bf RubbleSim: A Photorealistic Structural Collapse Simulator for \\Confined Space Mapping}
\author{ Constantine Frost$^{1}$, Chad Council$^{1}$, Margaret McGuinness$^{2}$, and Nathaniel Hanson$^{1}$
\thanks{$^{1}$Lincoln Laboratory, Massachusetts Institute of Technology, Lexington, Massachusetts, USA}%
\thanks{$^{2}$University of Notre Dame, Notre Dame, Indiana, USA}%
\thanks{Correspondence: {\tt\footnotesize nathaniel.hanson@ll.mit.edu}}%
\thanks{
DISTRIBUTION STATEMENT A. Approved for public release. Distribution is unlimited.
This material is based upon work supported by the Department of the Air Force under Air Force Contract No. FA8702-15-D-0001 or FA8702-25-D-B002. Any opinions, findings, conclusions or recommendations expressed in this material are those of the author(s) and do not necessarily reflect the views of the Department of the Air Force.}%
\thanks{© 2025 Massachusetts Institute of Technology.}%
\thanks{Subject to FAR52.227-11 Patent Rights - Ownership by the contractor (May 2014) Delivered to the U.S. Government with Unlimited Rights, as defined in DFARS Part 252.227-7013 or 7014 (Feb 2014). Notwithstanding any copyright notice, U.S. Government rights in this work are defined by DFARS 252.227-7013 or DFARS 252.227-7014 as detailed above. Use of this work other than as specifically authorized by the U.S. Government may violate any copyrights that exist in this work.}
}

\begin{document}

\maketitle
\thispagestyle{empty}
\pagestyle{empty}

\begin{abstract}
Despite well-reported instances of robots being used in disaster response, there is scant published data on the internal composition of the void spaces within structural collapse incidents. Data collected during these incidents is mired in legal constraints, as ownership is often tied to the responding agencies, with little hope of public release for research. While engineered rubble piles are used for training, these sites are also reluctant to release information about their proprietary training grounds. To overcome this access challenge, we present RubbleSim -- an open-source, reconfigurable simulator for photorealistic void space exploration. The design of the simulation assets is directly informed by visits to numerous training rubble sites at differing levels of complexity. The simulator is implemented in Unity with multi-operating system support. The simulation uses a physics-based approach to build stochastic rubble piles, allowing for rapid iteration between simulation worlds while retaining absolute knowledge of the ground truth. Using RubbleSim, we apply a state-of-the-art structure-from-motion algorithm to illustrate how perception performance degrades under challenging visual conditions inside the emulated void spaces. Pre-built binaries and source code to implement are available online: \url{https://github.com/mit-ll/rubble_pile_simulator}.

\end{abstract}

\section{Introduction \& Related Work}
When disaster strikes and buildings collapse, a race against time begins to rescue victims from beneath the rubble. Fortunately, severe building collapses are rare, but this rarity makes gaining experience in real-world incidents a challenge \cite{Murphy2016}. Urban search and rescue (USAR) teams train on engineered or simulated building collapses to hone the technical skills needed to conduct safe and efficient extractions of victims from the rubble piles. These training sites are typically built from reinforced concrete in strategic locations to ensure safe entry points for crews to access void spaces. These piles are also populated with other ``props''---elements such as vehicles, bricks, and metal---to increase mobility challenges for the humans and canines who train on the piles.

While these piles are complex enough to allow for multiple  scenarios for human operators, many practical considerations make them non-ideal for autonomous systems. Chiefly, they remain dangerous environments that require training and supervision to access. Although they are designed to prevent secondary collapses, they are still replete with open voids and rapidly changing surface conditions. These environments constitute some of the most challenging situations that robots might ever encounter. The repeated loss of robots used to collect training data to enable motion planning or perception algorithms for use during structural surveys~\cite{Ambe2016Kumamoto} or rescue operations~\cite{murphy2004trial} impedes the development of robots to assist in these missions. As a consequence, prior efforts to understand the locations of void spaces within 3D structures have used aerial platforms \cite{rao2022analysis, hu2022seeing} to observe collapse piles remotely rather than entering the structure.

\begin{figure}[t]
    \centering
    \vspace{0.5em}
    \includegraphics[width=\linewidth]{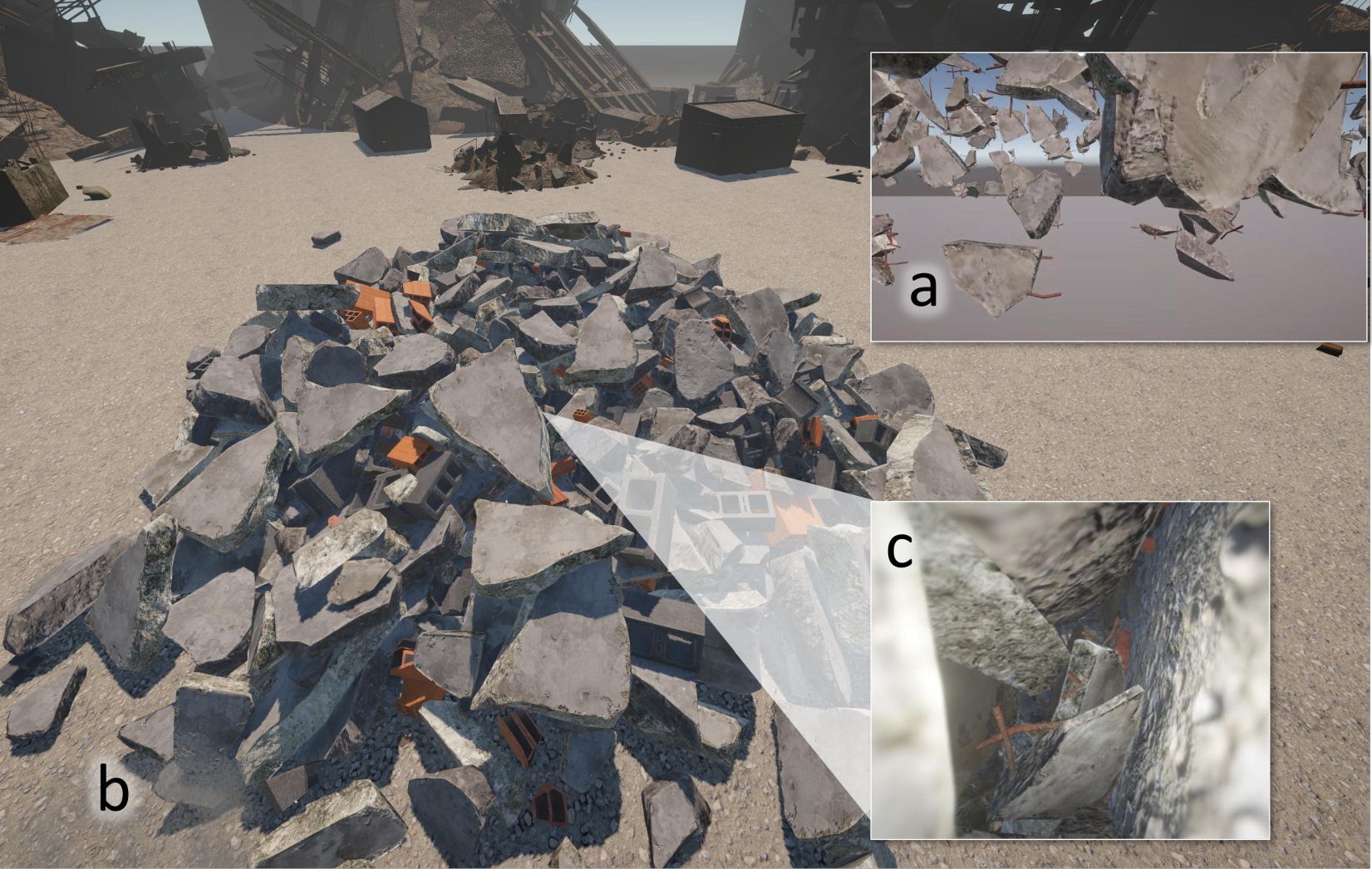}
    \caption{\textbf{RubbleSim Render of Collapsed Structure.} (a) Physics-based deposition of structural elements.  (b) Full pile rendered with clear conditions and lighting. (c) View from within a void space beneath the surface of the pile.}
    \label{fig:rubble_sim_teaser}
    \vspace{-1.5em}
\end{figure}
\begin{figure*}[t]
    \vspace{0.5em}
    \centering
    \includegraphics[width=\linewidth]{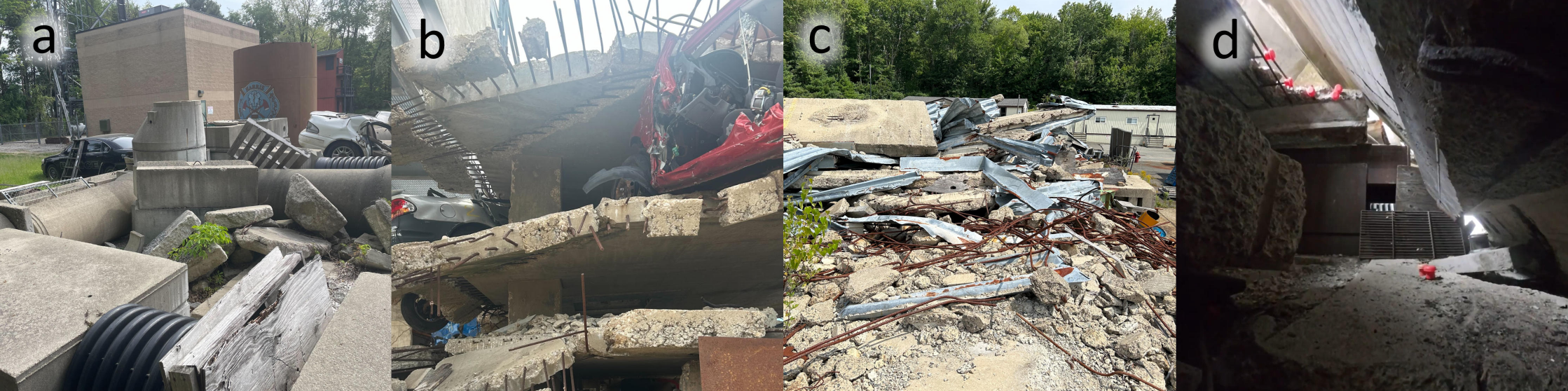}
    \caption{\textbf{Real-World Training Locations from Field Visits.} (a) Single layer training grounds with concrete culverts. (b) Partially-collapsed parking garage with crushed vehicles. (c) Full pancake collapse from concrete reinforced bridge with exposed guard rails and rebar. (d) Interior void space from within the pile in (c).}
    \label{fig:real_rubble_piles}
    \vspace{-1.5em}
\end{figure*}

Within the robotics research community, various groups have explored the use of mapping algorithms in confined spaces \cite{lim2008slam, tian2023visual}, but these algorithms have not been applied to rubble piles due to a lack of suitable evaluation data. In the literature, online mapping techniques, including ORB-SLAM2 \cite{mur2017orb}, NerfSLAM \cite{rosinol2023nerf}, and RTAB-Map \cite{labbe2019rtab}, have been widely employed for both indoor and outdoor mapping. Offline methods like Structure from Motion (SfM) \cite{ozyecsil2017survey} also create 3D meshes of an environment, albeit without localizing a robot in real-time. Other works have investigated the challenges of adapting algorithms for general confined spaces, such as in pipes \cite{aitken2021simultaneous, tian2023visual}, mines \cite{akbari2020intelligent, rogers2020test, ebadi2023present}, or industrial facilities \cite{helmberger2022hilti}. While several of these exemplar environments have similar aspects to those of collapsed structures, it is unknown how these general classes of algorithms will handle the dim, inconsistent lighting, the general lack of distinct visual features, and the short working distance from the scene to the camera --- all hallmarks of search and rescue robotics.

Given the trend towards data-intensive deep learning and mapping algorithms in robotics \cite{mokssit2023deep}, there is an unmet need for a simulator to emulate the void spaces of collapsed structures. In this work, we present a first-of-its-kind simulator to create rubble piles akin to those located at USAR training sites. RubbleSim is built on top of the popular Unity cross-platform game engine. In seconds, the simulator can create hundreds of variations of rubble piles of various depths and scales, with photorealistically rendered textures, lighting conditions, and environmental effects, as seen in Fig.~\ref{fig:rubble_sim_teaser}. The simulator integrates with both the ROS1 \cite{quigley2009} and the ROS2 \cite{doi:10.1126/scirobotics.abm6074} middleware, exposing void space environments to a wide variety of open-source robotics work.
This work provides the following contributions to the research community:
\begin{itemize}
    \item Pre-compiled, cross-platform simulator environments for rapidly generating rubble piles.
    \item Open-source framework with an extensible asset collection to increase the complexity of the environments.
    \item Evaluation of the performance of state-of-the-art structure from motion (SfM) algorithm within the void spaces of the simulated rubble piles.
\end{itemize}

\section{Simulator Environment}
\begin{table}[t]
    \centering
    \caption{\bf{RubbleSim Configuration Parameters.}}
    \label{tab:sim_params}
    \renewcommand{\arraystretch}{1.0}
    \setlength{\tabcolsep}{2.5pt}
    \begin{tabular}{p{0.30\columnwidth} p{0.65\columnwidth}}
        \toprule
        \textbf{Parameter} & \textbf{Description} \\
        \midrule
        \multicolumn{2}{l}{\textbf{Pile Spawn Arguments}} \\
        \midrule
        \texttt{spawnposx/y/z} & Center position of spawn volume \\
        \texttt{spawnboundx/y/z} & Volume scale along each axis  \\
        \texttt{numlayers} & Number of layers in pile \\
        \texttt{numobjs} & Objects per layer \\
        \midrule
        \multicolumn{2}{l}{\textbf{Environmental Lighting}} \\
        \midrule
        \texttt{setlightrot} & Enable fixed light rotation (else random) \\
        \texttt{lighttype} & Enum for spot/directional/point \\
        \texttt{lightintensity} & Light intensity \\
        \texttt{lightrot(x/y/z)} & Rotation about X, Y, Z \\
        \texttt{lightpos(x/y/z)} & Position along X, Y, Z \\
        \midrule
        \multicolumn{2}{l}{\textbf{Environmental Effects}} \\
        \midrule
        \texttt{fogdensity} & Baseline density for fog/smoke effect \\
        \texttt{fogintensity} & Fog/smoke noise variation \\
        \bottomrule
    \end{tabular}
    \vspace{-1.5em}
\end{table}

From engagements with teams at the local, state, and federal levels of USAR, we have visited a number of training grounds used to familiarize teams with confined space breach and ingress operations. Throughout these sites, depicted in Fig.~\ref{fig:real_rubble_piles}, there is a common element in the construction of the piles through the addition or rearrangement of props to increase the depth and complexity from a predefined set of rubble assets. These training piles evolve over time as props are reconfigured, allowing responders to practice navigation and victim extraction in new geometries without requiring major reconstruction. The process of physical reconfiguration inspired the modular, parametric approach that underpins RubbleSim’s world generation.

\begin{figure*}[tb]
    \vspace{0.5em}
    \centering
    \includegraphics[width=0.90\linewidth]{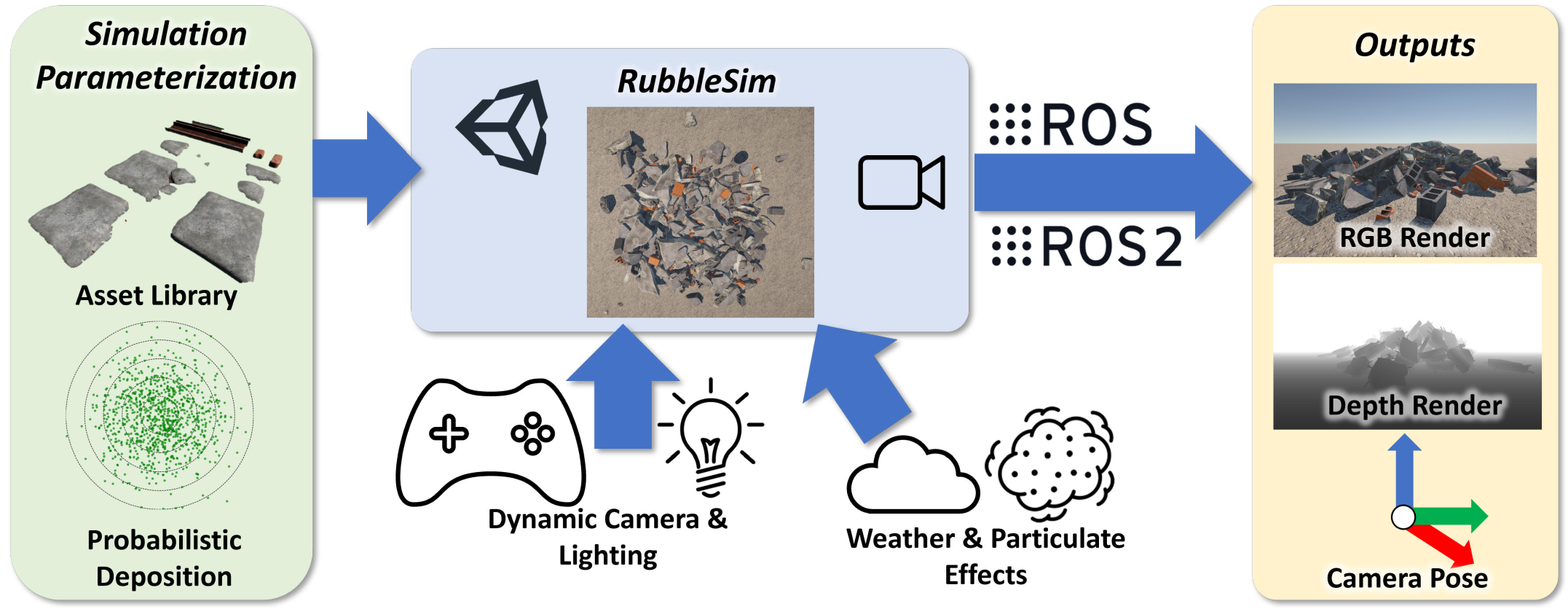}
    \caption{\textbf{RubbleSim Software Architecture.} Flowchart showing how RubbleSim creates rubble piles using a set of input simulation parameters. The created virtual world is then explored with a dynamic camera and point lighting conditions, while sensor inputs are marshaled through ROS1/ROS2.}
    \label{fig:rubble_sim_arch}
    \vspace{-1.5em}
\end{figure*}

RubbleSim seeks to emulate this construction paradigm within a flexible virtual environment. We selected Unity over other popular robotics simulation engines, such as NVIDIA Isaac Sim and Unreal Engine, for its mature ROS integration, efficient physics engine, and extensive library of 3D assets that are easily customized or procedurally varied. Unity’s ecosystem also supports real-time lighting models, stochastic environment generation, and GPU-accelerated rendering pipelines, all of which make it highly extensible for increasingly complex or photorealistic simulations. This flexibility allows RubbleSim to bridge the gap between low-fidelity physics-based testing and high-fidelity perceptual benchmarking.

 The development of this simulator was guided by the need to understand the geometries of spaces that robots working in USAR tasks would encounter and to evaluate how camera placement, lighting conditions, and surface geometry affect mapping and perception pipelines. RubbleSim, therefore, serves both as a visualization tool and as a data engine for algorithmic benchmarking. Command line arguments control the number of objects, size, and environmental conditions of the generated pile, allowing for reproducibility and batch generation of training datasets (Table~\ref{tab:sim_params}). Environmental parameters include pile footprint, height, asset number, lighting intensity, and particle effects. An overview of the simulator architecture, controls, and outputs is shown in Fig.~\ref{fig:rubble_sim_arch}.

\subsection{Collapse Pile Aggregation}
The process of building a pile begins with a predefined set of 3D assets that approximate common debris found in collapsed structures. From these predefined sets, users may specify the likelihood of introducing an asset into the pile through weighted probabilities. For instance, in a rubble pile simulating a parking garage, the prevalence of crushed vehicles should be sparse compared to the amount of concrete slabs and steel rebar. Similarly, a residential collapse can include drywall, wooden beams, and furniture. The file structure within the open-source code allows users to expand the simulator by adding new 3D assets with defined collision boxes and physical properties, such as density and friction. The current simulator comes bundled with a variety of concrete forms, cinder blocks, and bricks to create a number of large, medium, and small voids and entry holes into these spaces, each with a randomized pose and orientation.

\begin{figure*}[t]
    \vspace{0.5em}
    \centering
    \includegraphics[width=\linewidth]{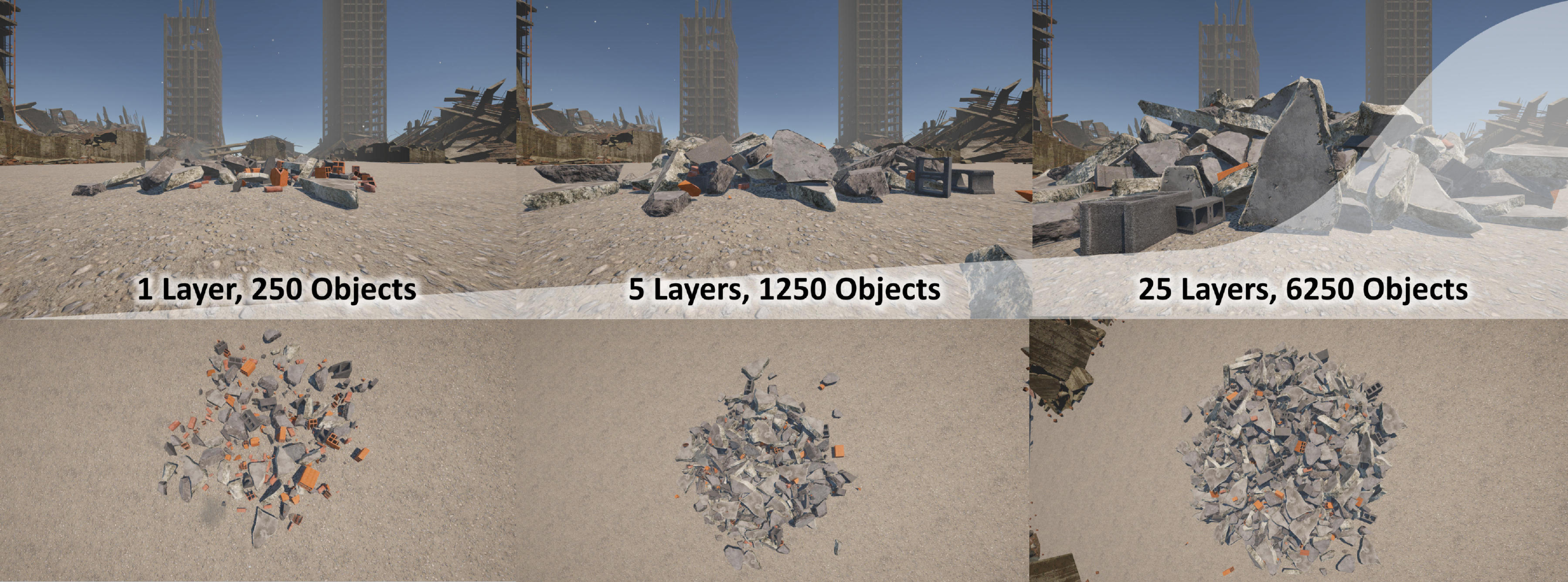}
    \caption{\textbf{Incremental Complexity of Rubble Piles.} RubbleSim exposes an interface to a number of components per stochastic deposition cycle. Increasing the number of layers increases both the footprint size and the total height of the rubble pile, with a greater chance of void space creation.}
    \label{fig:pile_growth}
    \vspace{-1.5em}
\end{figure*}

Assets are spawned on a series of random planes above the ground, at locations drawn from a probability distribution to create a position $\mathbf{P} (x,y,z)$ and a pose represented by a quaternion $\mathbf{q}$. By default, a uniform distribution is used; however, alternative distributions can be utilized to control the regions in which assets accumulate, thereby mimicking different collapse modes. After all assets have been spawned, they are subjected to Unity’s rigid-body physics and allowed to settle under gravity until a static configuration is found. The simulation proceeds until the system reaches equilibrium, capturing a chaotic yet stable stacking of debris. The process of spawning an environment averages less than two seconds and is replicable by setting a fixed seed value, thereby enabling consistent testing across trials. Fig.~\ref{fig:pile_growth} depicts the process of incrementally building the collapse pile. Once the pile has settled, it is exportable as a static 3D asset (\texttt{.stl} format).

\subsection{Camera Model \& Motion}
The accumulated pile is explorable using a camera model inspired by serpentine~\cite{whitman2018snake} and growing~\cite{mcfarland2024field} robots. The virtual camera can be moved manually, via keyboard or joystick teleoperation, or along scripted trajectories to simulate exploratory motion through narrow, irregular passages. The camera pose is instantaneously controlled in roll, pitch, and yaw, enabling precise navigation even in highly confined geometries. Once a directional vector is established, the camera is able to move forward or backward along its optical axis at adjustable speeds, allowing for slow inspection or rapid coverage. This motion model mirrors that of a head-mounted sensor on an extending robot tip, providing a relevant proxy for field operations.

\subsection{Perceptual Data Feeds}
As the camera moves through the environment, the RGB scene is captured, rendered, and published via a ROS1/ROS2 publisher as a 1024 $\times$ 1024 pixel image. The same field of view is also rendered as a registered depth image. Both the color and depth frames can be recorded and saved to rosbag files for later replay or training purposes. The position and rotation of the camera are jointly published as a \texttt{Pose} message to provide absolute global localization of the sensor within the scene—a necessary step for SLAM-based position tracking and evaluation of mapping accuracy.

\subsection{Environmental Effects}
\begin{figure}[t!t]
    \centering
    \includegraphics[width=\linewidth]{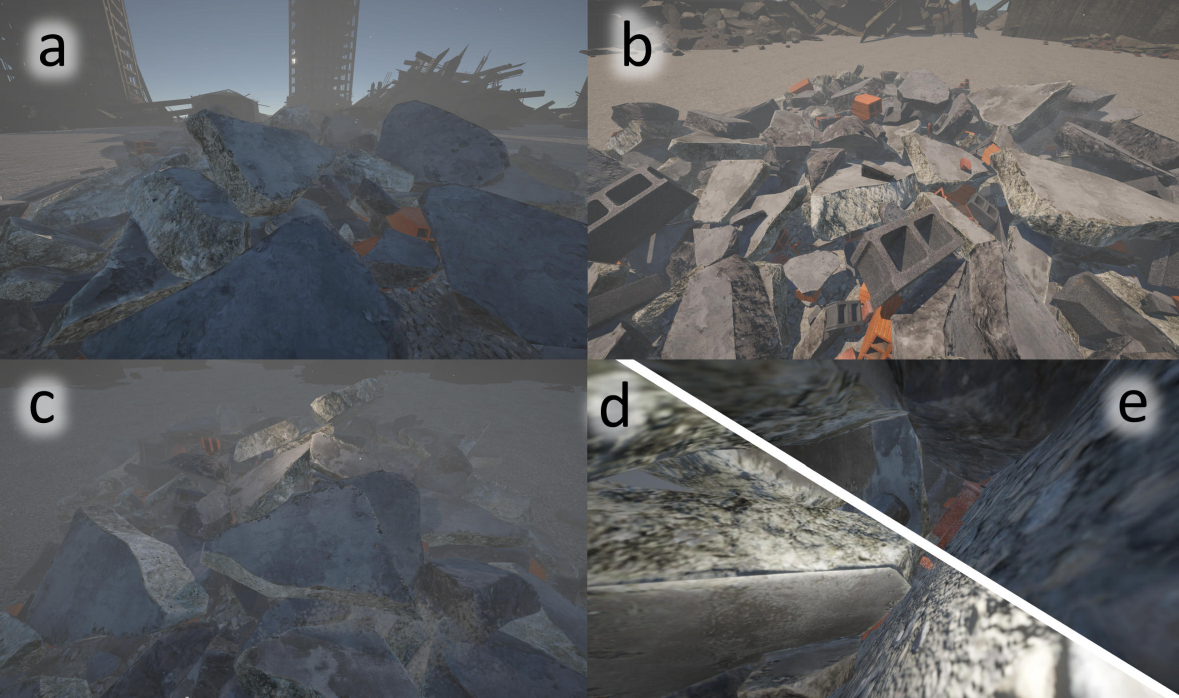}
    \caption{\textbf{RubbleSim Environmental Effects}. (a) Low-light on rubble pile. (b) Bright overhead illumination on rubble pile. (c) Momentary particulate effects simulating dust cloud and static environment fog. (d) Interior view of the rubble pile with variable lighting activated. (e) The same environment as (d) without added lighting.}
    \label{fig:sim_effects}
    \vspace{-1.0em}
\end{figure}

Previous uses of USAR robots have indicated a need for robust perception through airborne particulates, such as dust and smoke \cite{murphy2009mobile}. We add the ability to cover the environment with a layer of stochastic smoke or fog simulant that obscures visual features as a function of their distance from the camera. Additionally, we introduce perceptual noise in the form of periodic dust plumes on the pile. Fig~\ref{fig:sim_effects} shows examples of both the environmental effects and lighting controls.

The simulation’s global lighting system provides configurable control over the illumination conditions within the virtual collapse environment. Users can select the light type (spot, directional, or point), specify its intensity, and fix or randomize its orientation and position in 3D space. This flexibility allows the scene to emulate a wide range of visual conditions—from directional sunlight filtering through debris openings to a targeted spotlight illuminating the surface pile. This allows for an emulation of nearly all times of day.

The interior of the voids, beyond the initial entry points, quickly becomes dimly lit. We provide an optional point light co-located with the camera, which has a controllable beam intensity. The light model emulates ring lights or LED clusters typically found on small exploratory robots and exposes the same perceptual challenges faced by physical systems, including glare, specular reflections, and rapid intensity falloff with distance. These adjustable illumination parameters make RubbleSim particularly useful for testing low-light sensing approaches.

\section{Demonstration}
To assess the value of RubbleSim as a benchmarking environment and highlight the challenges that structural collapse poses to state-of-the-art mapping algorithms, we constructed a 15-layer rubble pile with 3,750 individual assets, including concrete slabs, rebar clusters, and cinder blocks. The resulting environment measured roughly $10~\mathrm{m} \times 10~\mathrm{m}$. With this pile, we collected data from three camera trajectories: a surface trajectory moving across the top of the pile, a subsurface trajectory into a void space without added lighting, and the same subsurface trajectory but with added lighting. For each trial, we downsampled synchronized RGB, depth, and ground-truth pose data to 30~Hz. In each evaluation, the lighting was direct and overhead.

\begin{figure}[tb]
    \centering
    \includegraphics[width=\linewidth]{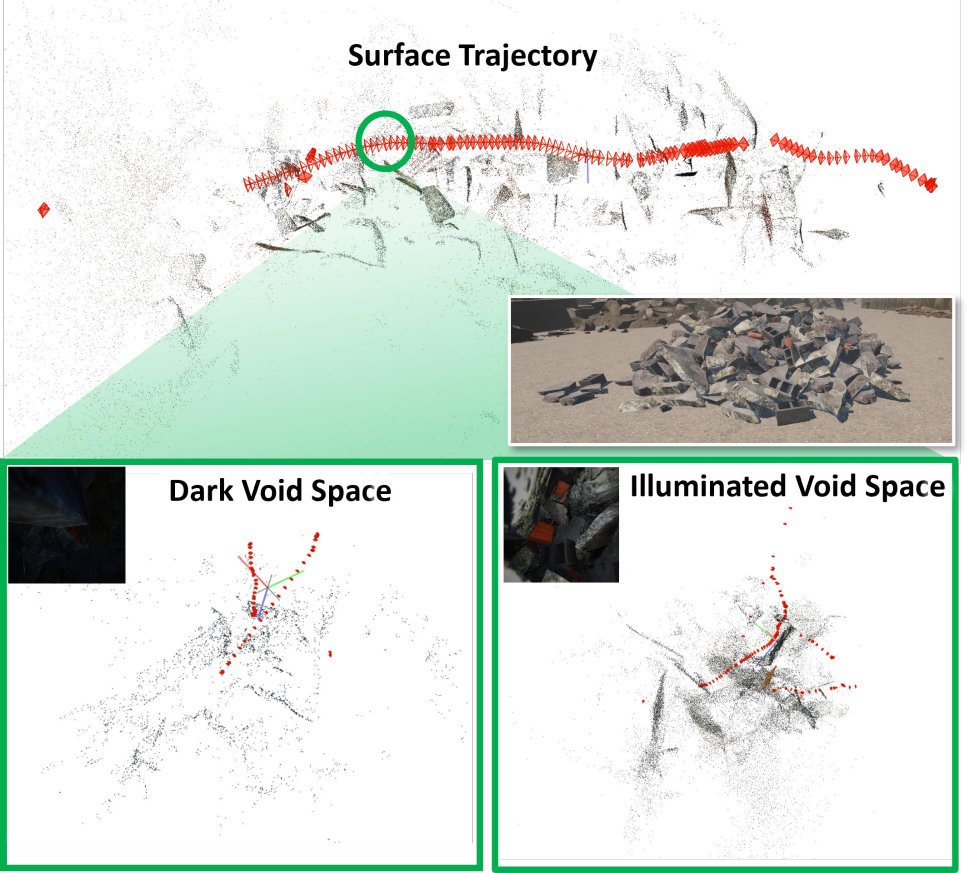}
    \caption{\textbf{Structure from Motion Trajectory and Results}. (Top) Sparse reconstruction of the rubble pile surface using COLMAP. The green circle indicates the location of the void used for subterranean exploration. (Inset) Rendered RGB view of the pile. (Bottom) Sparse mesh reconstructions for explorations of the void space with ambient (dark) and active illumination. Images in top-left corner of each show sample images from trajectory. Red markers are the COLMAP calculated camera perspectives for images.}
    \label{fig:sfm_qualitative}
    \vspace{-1.0em}
\end{figure}

With these trajectories, we utilize the COLMAP \cite{schoenberger2016sfm, schoenberger2016mvs} SfM pipeline to generate a mesh reconstruction of the pile and the camera viewpoints. We generate a dense surface using Poisson reconstruction \cite{kazhdan2013screened}. All experiments were performed on a PC with an Intel~i9 CPU and an RTX~4090 GPU, using ROS Noetic to output the images into a ROS bag. We report metrics related to the reconstruction in Table~\ref{tab:sfm_results}. \textit{Sequence Length} indicates the number of images in the trajectory sequences. \textit{Track Segments} counts the number of segments into which the trajectory is divided by COLMAP; this value should ideally be 1, as each trajectory represents a smooth, contiguous path over or through the pile. \textit{\% on Track} is the percentage of frames that match the trajectory sequence; ideally, this value should be 100\%. \textit{Points} is the number of image keypoints in the sparse reconstruction. Here, a greater number of keypoints is desirable. Fig.~\ref{fig:sfm_qualitative} shows a qualitative evaluation of the trajectories within the void spaces.

As shown in Table~\ref{tab:sfm_results}, COLMAP performs best when used on the exterior of the pile. The majority of the camera frame positions align with the actual trajectory, except for a few errant frames. The exploration of the void space with only ambient illumination (dark) shows a significant break in the trajectory, as the reconstruction fails under similarly low-texture surfaces. These results demonstrate that the standard SfM pipeline, though highly effective in well-lit outdoor scenes, degrades severely in confined and texture-poor conditions typical of collapse environments.

In particular, the reduced number of matched image keypoints (\textit{Points}) in the dark void sequence indicates that both visual contrast and parallax are insufficient for reliable feature triangulation. This limits not only point cloud density but also global consistency, as seen from the increase in trajectory breaks (\textit{Track Segments}). Illumination enhancement algorithms, such as CLAHE \cite{reza2004realization}, might help generate better keypoints to use in the reconstruction; however, they would need to be applied adaptively, as excessive contrast amplification could amplify sensor noise or motion blur present in low-light imagery with long exposure times.

\begin{table}[t]
    \centering
    \caption{\textbf{Structure from Motion (SfM) benchmarking results in RubbleSim.}}
    \label{tab:sfm_results}
    \renewcommand{\arraystretch}{1.0}
    \setlength{\tabcolsep}{3pt}
    \begin{tabular}{p{0.20\columnwidth} p{0.16\columnwidth} p{0.18\columnwidth} p{0.18\columnwidth} p{0.12\columnwidth}}
    \toprule
    \textbf{Dataset} & \textbf{Sequence Length} & \textbf{Track Segments} & \textbf{\% On Track} & \textbf{Points} \\
    \midrule
    Exterior & 126 & 1 & 92.6 & 79885 \\
    Void (Dark) & 88 & 2 & 65.9 & 4776 \\
    Void (Light) & 157 & 3 & 67.5 & 40521\\
    \bottomrule
    \end{tabular}
    \vspace{-1.5em}
\end{table}

For the illuminated void trial, reconstruction completeness improves substantially, with an order-of-magnitude increase in the number of observed keypoints (\textit{Points}) compared to the dark void trial. Nonetheless, trajectory fragmentation persists, suggesting that lighting alone is insufficient to fully mitigate the geometric and photometric challenges of subsurface environments. Even with the improved illumination, the variable geometry of the void spaces leads to moments of over-exposure, where parts of the scene are washed out, resulting in missed correspondences between subsequent images. This highlights the need for variable illumination to handle the changing working distances of the camera. These findings suggest that future reconstruction pipelines for subterranean or collapsed environments may benefit from active perception strategies, such as variable-gain illumination, structured light projection, or multimodal sensing (e.g., RGB+depth fusion), to maintain reliable localization and surface recovery across diverse lighting conditions.

Overall, these experiments demonstrate that RubbleSim effectively reproduces the degradation patterns observed in real-world USAR environments, providing a controlled means to benchmark algorithmic robustness under progressively challenging optical conditions.

\section{Conclusion and Future Work}
In this paper, we introduced RubbleSim, a photorealistic simulator designed to advance robotic perception and simulation-based research for urban search and rescue (USAR) applications. The presented results demonstrate RubbleSim’s ability to reproduce common SLAM degradation patterns and generate meaningful datasets for benchmarking mapping and perception algorithms under controlled yet realistic conditions.

While effective for perception and mapping studies, the current implementation omits several physical effects that are critical for locomotion and interaction modeling. Frictional contact between robots and rubble surfaces is not yet implemented, and the dust particle system remains purely visual, without affecting sensor noise or depth estimation. Future work will further incorporate Unity’s weather effects, improving physical realism and perceptual fidelity.

In addition, following feedback from the community, we plan to introduce more realistic camera artifacts, including motion blur from rapid movement and dynamic exposure adaptation in low-light environments, which are known to challenge visual perception algorithms. These extensions will complement ongoing work to integrate an event-camera model based on~\cite{rebecq18corl}, enabling multimodal sensing with RGB, depth, and event data, which is known to be particularly beneficial in low-light conditions~\cite{huang2023event}. Further integration with Unity ML-Agents~\cite{juliani2020} and the ROS Navigation Stack~\cite{macenski2020marathon2} will enable reinforcement learning, autonomous navigation, and motion planning research in confined environments.

Ultimately, RubbleSim is envisioned as an open, photorealistic testbed for evaluating perception, mapping, and autonomy algorithms in standardized rubble scenarios without risking the loss of robots during algorithm development. By progressively improving physical realism, environmental variability, and sensing modalities, we aim for RubbleSim to serve as a unifying platform bridging academic research and applied USAR robotics development.

\section*{Acknowledgments}

The authors are grateful to the members of Massachusetts Task Force 1 (MA-TF1), Ohio Task Force 1 (OH-TF1), New York City Fire Department (FDNY), Clay Fire Territory, and the Texas A\&M Engineering Extension Service (TEEX) Disaster City team for facilitating access to their training sites and providing insightful conversations to guide the development of this simulator.

\bibliographystyle{IEEEtran} 
\bibliography{references}

\addtolength{\textheight}{-12cm}   
\end{document}